\documentclass[letterpaper, 10 pt, journal, twoside]{ieeetran}

\usepackage{mathtools} 
\usepackage{amssymb}  
\usepackage{cite}
\usepackage{booktabs}
\usepackage[subnum]{cases}
\usepackage{balance}
\usepackage{color}
\usepackage{tabularx}
\usepackage{graphicx,dblfloatfix}
\usepackage[caption=false, font=footnotesize]{subfig}
\usepackage{listings}
\usepackage{textcomp}
\usepackage{url}
\usepackage{multirow}
\usepackage{algorithm}
\usepackage[noend]{algpseudocode}
\usepackage{ragged2e}
\newcolumntype{Y}{>{\RaggedRight\arraybackslash}X}

\usepackage{bm}
\usepackage[colorinlistoftodos]{todonotes}
\usepackage{hyperref}
\usepackage[normalem]{ulem}
\usepackage{siunitx} 
\usepackage{csquotes} 
\usepackage{xparse} 
\usepackage{soul} 
\usepackage{soulutf8} 
\soulregister\cite7 
\soulregister\autoref7 
\soulregister\figref7
\soulregister\ref7
\soulregister\eqref7
\soulregister\ac7 

\usepackage[printonlyused]{acronym}

\acrodef{RSL}{Robotic Systems Lab}
\acrodef{ASL}{Autonomous Systems Lab}
\acrodef{ETHZ}{Swiss Federal Institute of Technology Zurich}
\acrodef{ZMP}{zero-moment point}
\acrodef{CoM}{center of mass}
\acrodef{EoM}{equations of motion}
\acrodef{DoF}{degree of freedom}
\acrodef{MPC}{model predictive control}
\acrodef{QP}{quadratic program}
\acrodef{LQR}{linear-quadratic regulator}
\acrodef{TO}{trajectory optimizer}
\acrodef{LHS}{left-hand side}
\acrodef{RHS}{right-hand side}
\acrodef{WBC}{whole-body control}
\acrodef{NLP}{nonlinear program}
\acrodef{PD}{proportional-derivative}
\acrodef{PI}{proportional-integral}
\acrodef{PID}{proportional-integral-derivative}
\acrodef{ZOH}{zero-order hold}
\acrodef{ROS}{Robot Operating System}
\acrodef{IMU}{Inertial Measurement Unit}
\acrodef{LoS}{line of support}
\acrodef{DoF}{degree of freedom}
\acrodef{DoFs}{degrees of freedom}
\acrodef{KF}{Kalman Filter}
\acrodef{UKF}{Unscented Kalman Filtering}
\acrodef{PF}{Particle Filter}
\acrodef{w.r.t.}{with respect to}
\acrodef{DARE}{discrete time algebraic Ricatti equation}
\acrodef{CARE}{continuous time algebraic Ricatti equation}
\acrodef{CAD}{computer-aided design} 

\renewcommand{\vec}{\bm}
\newcommand{\mat}{\bm}

\newcommand{\skewmat}[1]{\left[ #1 \right]_{\times}}
\newcommand{\figref}{Fig.~\ref}
\newcommand\sectionspace{0}
\newcommand\subsectionspace{0}

\title{LQR-Assisted Whole-Body Control of a Wheeled Bipedal Robot with Kinematic Loops}

\author{Victor~Klemm, Alessandro~Morra, Lionel~Gulich, Dominik~Mannhart, \\David~Rohr, Mina~Kamel, Yvain~de~Viragh, and Roland~Siegwart 
\thanks{Manuscript received: September, 10, 2019; Revised: December, 17, 2019; Accepted: January, 19, 2020.}
\thanks{This paper was recommended for publication by Editor Nikos Tsagarakis upon evaluation of the Associate Editor and Reviewers' comments.}%
\thanks{Yvain~de~Viragh is with the CRL, Department of Computer Science, ETH Z\"urich, 8092 Z\"urich, Switzerland, email: {\tt\small yvaind@ethz.ch}.}%
\thanks{All other authors are with the ASL, ETH Z\"urich, 8092 Z\"urich, Switzerland, email: {\tt\small vklemm@ethz.ch}, {\tt\small morraa@ethz.ch}, {\tt\small lgulich@ethz.ch}, {\tt\small dominikm@ethz.ch}, {\tt\small rohrd@ethz.ch}, {\tt\small fmina@ethz.ch}, {\tt\small rsiegwart@ethz.ch}.}%
\thanks{Digital Object Identifier (DOI): see top of this page.}
}

\begin{document}
\markboth{IEEE Robotics and Automation Letters. Preprint Version. Accepted January, 2020}{Klemm \MakeLowercase{\textit{et al.}}: LQR-Assisted Whole-Body Control of a Wheeled Bipedal Robot with Kinematic Loops}
\maketitle

\begin{abstract}
We present a hierarchical whole-body controller leveraging the full rigid body dynamics of the wheeled bipedal robot \emph{Ascento}. We derive closed-form expressions for the dynamics of its kinematic loops in a way that readily generalizes to more complex systems. The rolling constraint is incorporated using a compact analytic solution based on rotation matrices. The non-minimum phase balancing dynamics are accounted for by including a linear-quadratic regulator as a motion task. Robustness when driving curves is increased by regulating the lean angle as a function of the zero-moment point. The proposed controller is computationally lightweight and significantly extends the rough-terrain capabilities and robustness of the system, as we demonstrate in several experiments.
\end{abstract}

\begin{IEEEkeywords}
Legged Robots, Wheeled Robots, Parallel Robots, Dynamics, Robust/Adaptive Control of Robotic Systems
\end{IEEEkeywords}


\vspace{\sectionspace cm} \section{Introduction}
\label{sec:introduction}
\begin{figure}[!t]
	\centering
	\includegraphics[width=\columnwidth]{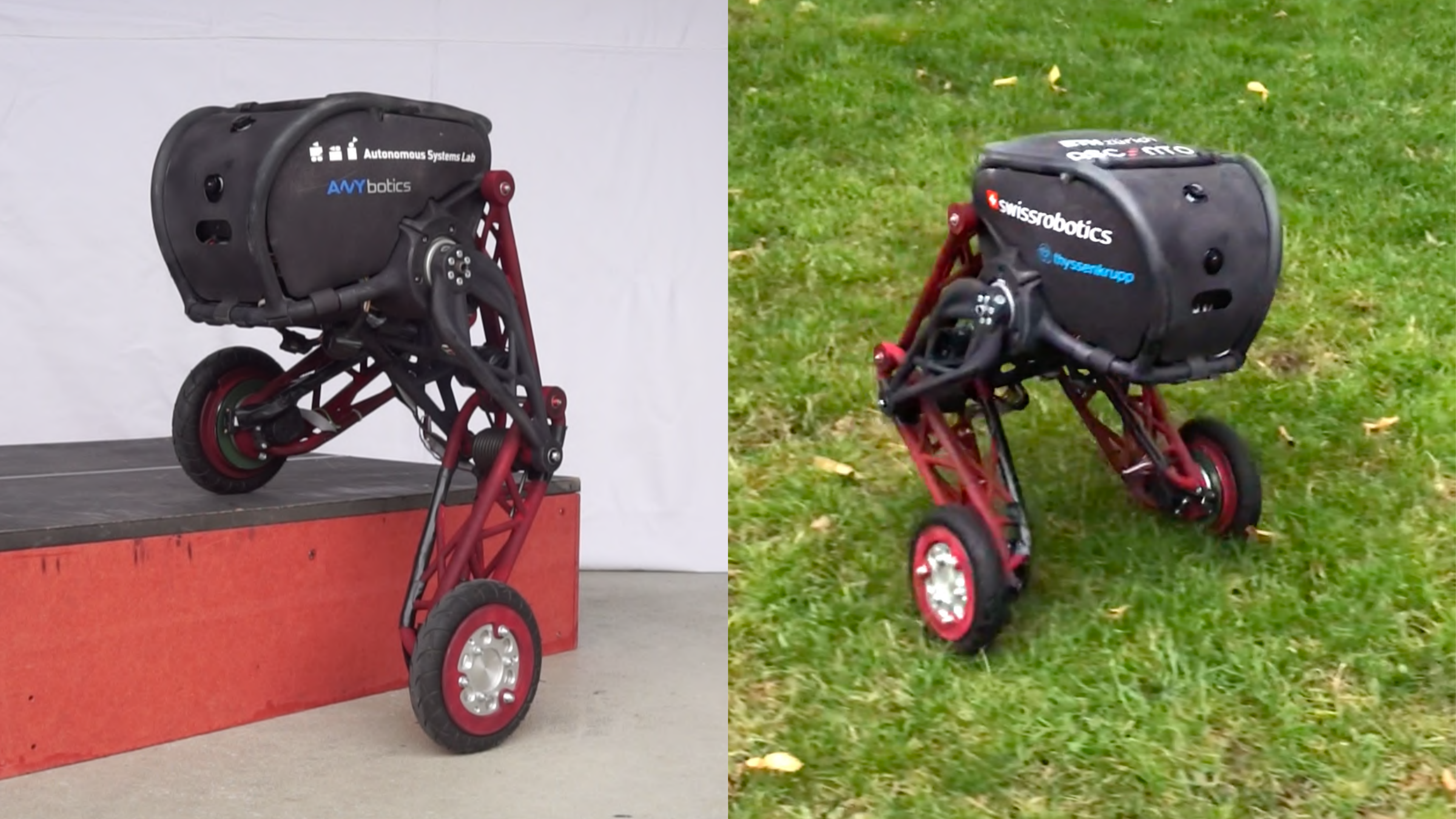}
    \vspace{-0.7cm}
	\caption{\emph{Left:} The \emph{Ascento} robot stabilizing while one wheel is placed on an elevated step of height \SI{0.20}{m}. The four-bar linkage of the left leg is nearly fully extended. \emph{Right:} The robot actively adjusts its legs when driving on uneven terrain, such as the grass slope shown.
	To see these and many more maneuvers in action we encourage the reader to watch the accompanying video: \url{https://youtu.be/nGu2odkB5ws}.}
	\vspace{-0.5cm}
	\label{fig:title}
\end{figure}

    \IEEEPARstart{F}{ast} and agile maneuverability is a key component for an efficient deployment of mobile ground robots. In this regard, wheeled-legged systems combine the best of two worlds -- they leverage both the speed and efficiency of wheels, and the ability of legs to overcome uneven terrain and obstacles. In recent years, wheeled bipedal robots have started to show the capabilities required for real-world applications~\cite{handle2}, while allowing for swift and cost-effective designs, requiring less actuators and being natively able to turn on spot.
The wheeled bipedal robot \emph{Ascento}\footnote{More information can be found on: \url{https://www.ascento.ethz.ch}.} presented in our previous work~\cite{ascento_paper} is capable of achieving many of the specifications required for typical applications. Being a parallel robot with a four-bar linkage, i.e. a kinematic loop in each of its legs, \emph{Ascento} only requires four actuators, two for driving the wheels and two for moving the legs. This reduces cost, weight, and mechanical complexity, which is desirable for inspection tasks as well as search and rescue applications.
As we have shown, a basic version of these tasks can already be completed with simplified, model-based control strategies~\cite{ascento_paper}. The new \ac{LQR}-assisted \ac{WBC} scheme proposed in this letter extends \emph{Ascento}'s capabilities to outdoor scenarios, rendering the robot more robust to disturbances by active compliance to uneven terrain, as shown in \figref{fig:title}.

\vspace{\subsectionspace cm} \subsection{Related Work}
\label{related_work}
Hierarchical inverse dynamics control (e.g.~\cite{hutter2014quadrupedal,herzog2014balancing}) and \ac{WBC} (e.g.~\cite{escande2014hierarchical, bellicos2016perception, kim2018computationally}) rapidly gained popularity over the last decade and have been applied to walking robots such as bipeds~\cite{sentis2006whole} and quadrupeds~\cite{bellicoso2017dynamic,winkler2015planning}. Recent works also showed successful deployment on a wheeled-legged quadruped~\cite{bjelonic2018keep,yvain_mt}.
However, to the best of the authors' knowledge, application of \ac{WBC} to stabilization of wheeled bipedal robots and their inherent non-minimumphase dynamics has not been shown before.\footnote{\emph{Boston Dynamic}'s wheeled bipedal robot \emph{Handle}~\cite{handle} has demonstrated impressive performance, but, unfortunately, little is known about the underlying control approaches.} In our previous work~\cite{ascento_paper}, we modeled the robot as a standard two-wheeled inverted pendulum, thereby completely neglecting leg dynamics. We improve this by rigorously treating the kinematic loops in the legs. Typically one approach of the following three is applied for modeling kinematic loops: 
1) The system dynamics are derived from an explicit formulation of the kinematics, which is trivial for linkages with a simple geometry~\cite{siciliano2010robotics}, but becomes considerably more involved for loops with irregular link lengths~\cite{mina}, such as the one of \emph{Ascento}. Further, this approach is only directly applicable to linkages without kinematic inversions in their operating space~\cite{hartenberg:kinematic_linkages}.
2) The system dynamics can be be found by purely numerical techniques, such as the recursive Newton-Euler formulation described in \cite{newteul, khalil, khalil2004modeling}. However, this approach focuses rather on the simulation than on the derivation of a system's \ac{EoM}.
3) The loop is opened kinematically and closed by finding appropriate dynamic constraint forces~\cite{siciliano2016springer, iwamura2013method}.
We build on the third approach because it allows to derive closed-form solutions and can be applied to non-trivial systems.

\vspace{\subsectionspace cm} \subsection{Contribution}
\label{subsec:contribution}
Our main contributions can be summarized as follows:
\begin{itemize}
    \item Derivation of the full rigid body dynamics of a wheeled bipedal robot, with an emphasis on modeling kinematic loops (Section~\ref{subsec:loop_closure}).
    \item A compact and closed-form rolling constraint formulation using rotation matrices (Section~\ref{subsec:ground_contacts}).
    \item Synthesis of a \ac{WBC} scheme for control of such robots. Control of the non-minimum phase balancing dynamics is achieved by including an \ac{LQR} feedback law as a motion task. The robustness against tipping over when driving curves is increased by controlling the leaning angle such that the \ac{ZMP}~\cite{vukobratovic2004zero} is shifted towards the center of the \ac{LoS} (Section~\ref{sec:control}).
\end{itemize}
We demonstrate the performance of our \ac{WBC} scheme in Section~\ref{sec:results_and_discussion} and conclude by an outlook on future work in Section~\ref{sec:conclusion_and_outlook}.


\vspace{\sectionspace cm} \section{Modeling}
\label{sec:modeling}
    Since the \ac{WBC} introduced in Section~\ref{sec:control} is a model-based control technique, accurately modeling the system dynamics is key. It should be noted that the approaches presented in this section are generally applicable. For the sake of simplicity, we show them directly for the example of the \emph{Ascento} robot.

\vspace{\subsectionspace cm} \subsection{Coordinates and Conventions}
\label{coordinates_and_conventions}
\begin{figure}
	\centering
	\includegraphics[width=\columnwidth]{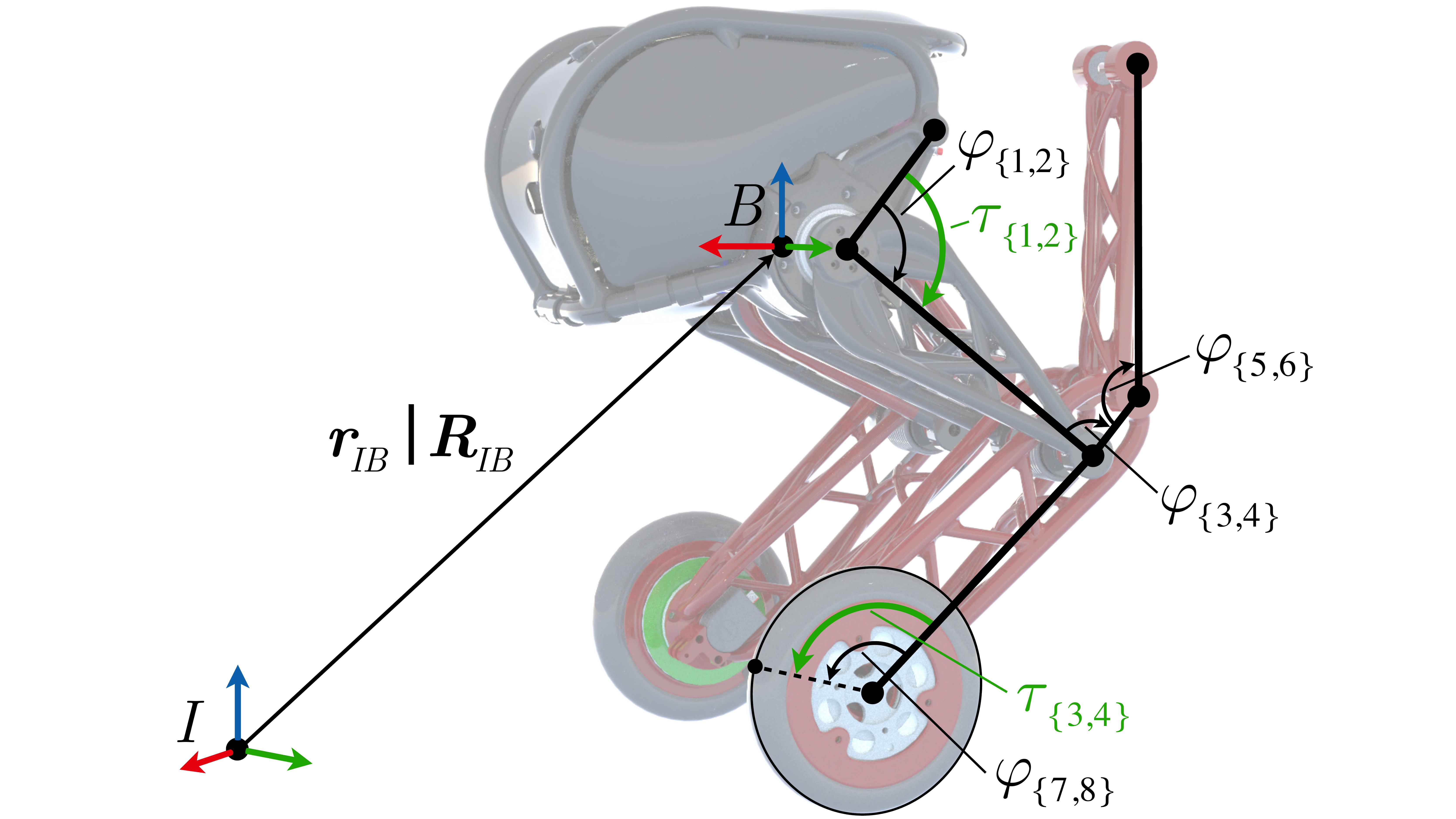}
    \vspace{-0.7cm}
	\caption{Generalized coordinates of the system with opened kinematic loops as introduced in \eqref{eq:gen_cor}. The joint angles and actuation torques are marked on the left leg only, and their indices denote the corresponding joint angle on the left- and right-hand side respectively: $\varphi_{\{l,r\}}$, $\tau_{\{l,r\}}$.}
	\vspace{-0.5cm}
	\label{fig:gen_cor}
\end{figure}

We define the generalized coordinates (\figref{fig:gen_cor}), velocities, accelerations, and actuation torques as
\begin{equation}
    \vec{q}=
        \begin{bmatrix}
        {}_{I}\vec{r}_{IB} \\
        \mat{R}_{IB} \\
        \varphi_1 \\
        \vdots \\
        \varphi_{n_{j}}
    \end{bmatrix}, \ \vec{u}=
    \begin{bmatrix}
        {}_{I}\vec{v}_{IB} \\
        {}_{I}\vec{\omega}_{IB} \\
        \dot{\varphi}_1 \\
        \vdots \\
        \dot{\varphi}_{n_{j}}
    \end{bmatrix}, \ \dot{\vec{u}}=
    \begin{bmatrix}
        {}_{I}\vec{a}_{IB} \\
        {}_{I}\dot{\vec{\omega}}_{IB} \\
        \ddot{\varphi}_1 \\
        \vdots \\
        \ddot{\varphi}_{n_{j}}
    \end{bmatrix}, \ \vec{\tau}=
        \begin{bmatrix}
            \tau_1 \\
            \tau_2 \\
            \tau_3 \\
            \tau_4 \\
    \end{bmatrix},
\label{eq:gen_cor}
\end{equation}
respectively, where $\vec{q} \in \mathbb{R}^{3} \times SO(3) \times \mathbb{R}^{n_{j}}$, $\vec{u} \in \mathbb{R}^{6+n_j}$, $\dot{\vec{u}} \in \mathbb{R}^{6+n_j}$ and $\vec{\tau} \in \mathbb{R}^{n_{\tau}}$. Thereby, ${}_{I}\vec{r}_{IB}$ describes the relative position vector of the frames in the right-hand subscript (i.e. from the inertial frame $I$ to the base frame $B$), represented in the frame of the left-hand subscript. Similar holds for the linear and angular velocities, and accelerations, i.e. ${}_{I}\vec{v}_{IB}$, ${}_{I}\vec{\omega}_{IB}$ and ${}_{I}\vec{a}_{IB}$, ${}_{I}\dot{\vec{\omega}}_{IB}$, respectively. The rotation matrix $\mat{R}_{IB}$ maps from coordinate representation in frame $B$ to frame $I$. Further, the scalars $\varphi_i$ represent the joint angles. In the following, we use superscript brackets (e.g. $\mat{J}^{(1,3)}$) to indicate specific rows of matrices and $\skewmat{\vec{\cdot}}$ for the skew-symmetric cross product matrix of a vector. We use the positional and rotational Jacobian convention
\begin{equation}
\label{eq:Jacobians}
\begin{bmatrix} {}_{I}\vec{v}_{IB} \\ {}_{I}\vec{\omega}_{IB} \end{bmatrix} = \begin{bmatrix} {}_{I}\mat{J}_{IB,\mathcal{P}} \\ {}_{I}\mat{J}_{IB,\mathcal{R}} \end{bmatrix} \vec{u}.
\end{equation}

\vspace{\subsectionspace cm} \subsection{Open-Loop Dynamics}
\label{open_loop_dynamics}
The unconstrained dynamics of the system with opened, kinematic loops can be formulated as
\begin{equation}
\label{eq:ol_eom}
\mat{M}(\vec{q}) \, \dot{\vec{u}} + \vec{b}(\vec{q}, \vec{u}) + \vec{g}(\vec{q}) + \vec{s}(\vec{q}) = \mat{S}^\top \vec{\tau},
\end{equation}
where $\mat{M}(\vec{q}) \in  \mathbb{R}^{n_{u} \times n_{u}}$ denotes the mass matrix, $\vec{b}(\vec{q}, \vec{u}) \in \mathbb{R}^{n_{u}}$ denotes the vector of Coriolis and centrifugal terms, and $\vec{g}(\vec{q}) \in \mathbb{R}^{n_{u}}$ is the vector of gravity terms.\footnote{These can, for instance, be calculated using the projected Newton-Euler equations, given the \ac{CoM} Jacobians and inertial parameters of each body.} Additionally, $\vec{s} \in \mathbb{R}^{n_{u}}$ accounts for the effect of two torsional springs in the knees of the robot. A linear-elastic spring law relating angular deflection and generated torque is assumed. The selection matrix $\mat{S} \in \mathbb{R}^{n_{\tau} \times n_{u}}$ selects on which generalized coordinates the actuation torques $\vec{\tau}$ are acting. In the following, the direct dependence on the generalized coordinates and velocities is omitted for brevity of notation.

\vspace{\subsectionspace cm} \subsection{Loop Closure}
\label{subsec:loop_closure}
The opened kinematic loop structure is dynamically closed by introducing loop closure forces $\tilde{\vec{F}}_L$ at the opened hinge points, as shown in Fig.~\ref{fig:loop}. In the following, firstly the derivations are performed exemplary for one loop only, and then applied to both of them. 
$\tilde{\vec{F}}_L$ can be interpreted as bearing forces with their respective reactions acting on the hinge points of the opened loop, $P$ and $Q$, and are directly added to \eqref{eq:ol_eom}:
\begin{figure}
	\centering
	\includegraphics[width=\columnwidth]{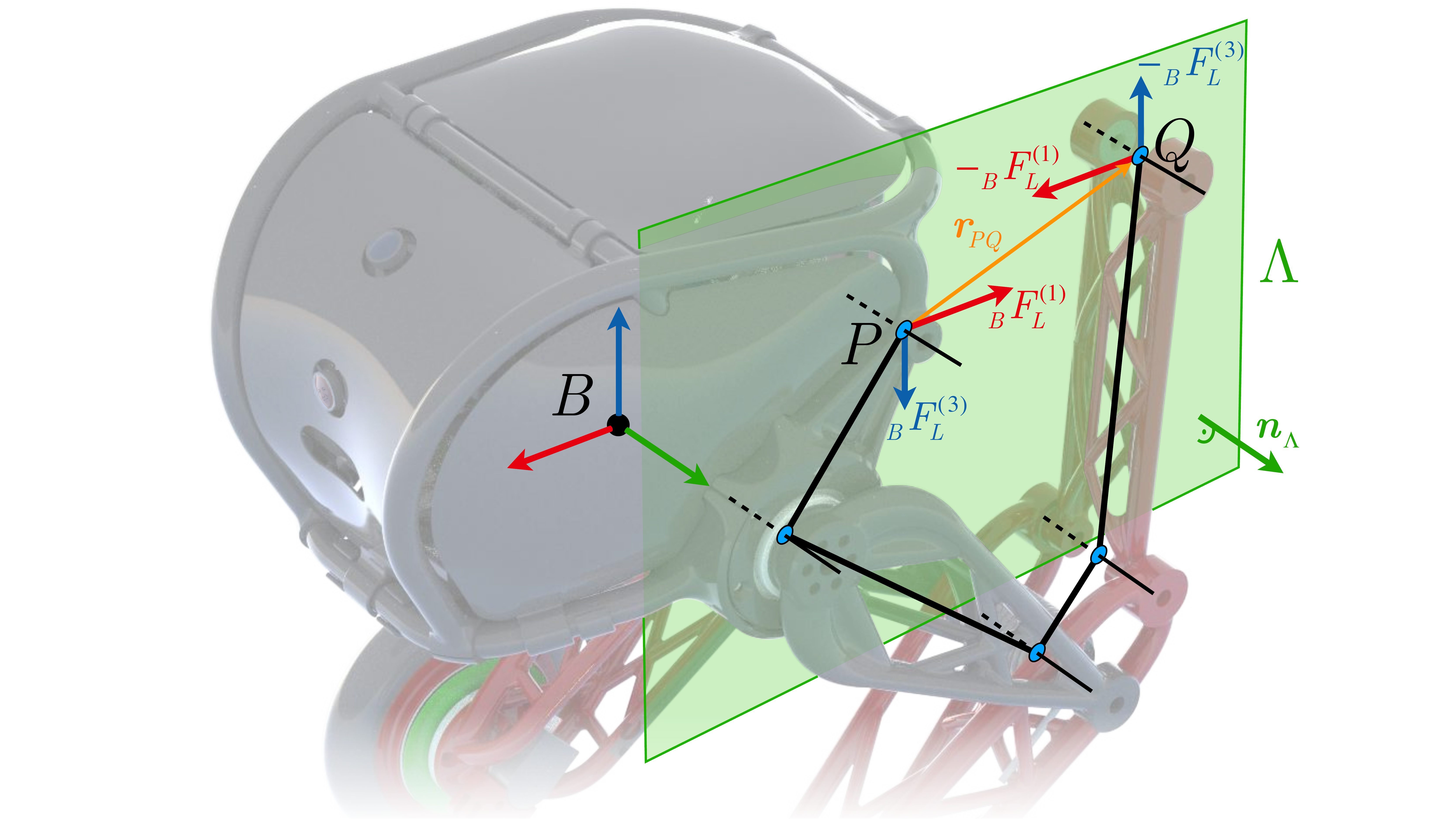}
    \vspace{-0.7cm}
	\caption{The opened kinematic loop (of the robot's left side). Indicated are the two hinge points $P$ and $Q$ where the loop has been opened. In order to close the loop, the two equal but opposite loop closure force pairs $\pm {}_{B}\vec{F}_{l}^{(1)}$ and $\pm {}_{B}\vec{F}_{l}^{(3)}$ acting in the loop-plane $\Lambda$ on $P$ and $Q$ must be found.}
	\vspace{-0.5cm}
	\label{fig:loop}
\end{figure}
\begin{equation}
\mat{M} \, \dot{\vec{u}} + \vec{b} + \vec{g} + \vec{s} + {}_{I}\mat{J}_{IP,\mathcal{P}}^\top \, {}_{I}\tilde{\vec{F}}_{L} - {}_{I}\mat{J}_{IQ,\mathcal{P}}^\top \, {}_{I}\tilde{\vec{F}}_{L} = \mat{S}^\top \vec{\tau}.
\end{equation}
It is to be noted that the loop closure forces should only act in the loop plane $\Lambda$, as its normal direction $\vec{n}_{\Lambda}$ is already constrained by the kinematics of the opened loop.\footnote{As loop closure forces in direction of $\vec{n}_{\Lambda}$ have no effect on the rigid mechanism, they can be of arbitrary size. This leads to rank-deficiency when calculating accelerations, as shown later in Section~\ref{subsec:equations_of_motion}.} Because the robot is a floating-base system, $\Lambda$ is changing over time. As the loops are fixed and aligned w.r.t. the base frame $B$, the in-plane components of the loop closure forces can readily be selected: They are directly the $x$- and $z$-components of their representation in the base frame,\footnote{It can be argued that the formulation becomes more general if these components are expressed in the frames attached to one of the opened hinge points $P$ or $Q$, but the formulation chosen here leads to simpler and more accessible expressions, depending directly on $\vec{q}$ and $\vec{u}$.} giving
\begin{equation}
\mat{M} \, \dot{\vec{u}} + \vec{b} + \vec{g} + \vec{s} + \underbrace{({}_{B}\mat{J}_{IP,\mathcal{P}}^{(1,3)} - {}_{B}\mat{J}_{IQ,\mathcal{P}}^{(1,3)})^{\top}}_{:=\tilde{\mat{J}}^\top_L} \, {}_{B}\tilde{\vec{F}}_{L}^{(1,3)} = \mat{S}^\top \vec{\tau}.
\end{equation}
This holds analogous for the left $(l)$ and right $(r)$ kinematic loop, resulting in similar expressions that can be stacked:
\begin{equation}
\label{eq:loop_closure_dynamics_compressed}
        \mat{M} \, \dot{\vec{u}} + \vec{b} + \vec{g} + \vec{s} +
        \underbrace{\begin{bmatrix}
        \tilde{\mat{J}}_{L_l}^\top & \tilde{\mat{J}}_{L_r}^\top
        \end{bmatrix}}_{:=\mat{J}_L^\top} \, \underbrace{\begin{bmatrix}
         {}_{B}\tilde{\vec{F}}_{L_l}^{(1,3)} \\  {}_{B}\tilde{\vec{F}}_{L_r}^{(1,3)}
        \end{bmatrix}}_{:=\vec{F}_L} =\mat{S}^{\top} \, \vec{\tau}.
\end{equation}
To determine the unknown loop closure forces, the position constraint $\vec{r}_{PQ} = \vec{0}$ is expressed at acceleration level. Similar to the loop closure forces, it must only be enforced in directions where there are still \ac{DoFs} to constrain, i.e. only in the loop plane $\Lambda$: This is done by the constraint projection ${}_{{B}}\vec{r}_{PQ}^{(1,3)} =  \vec{0}$.
To bring the positional constraint to the acceleration level, two-fold differentiation w.r.t. time is performed, taking care of differentiation in the potentially rotating base frame $B$:
\begin{align}
    {}_{B}\ddot{\vec{r}}_{PQ} =
    \frac{d^2}{dt^2}(\mat{R}_{BI} \, {}_{I}\vec{r}_{PQ}) \overset{(1,3)}{=} &\vec{0}, \\
    \frac{d}{dt}(\mat{R}_{BI} \, \skewmat{{}_{I}\vec{\omega}_{BI}} 
    \, {}_{{I}}\vec{r}_{PQ}
    +\mat{R}_{{BI}} \, {}_{I}\vec{v}_{PQ})
    \overset{(1,3)}{=} &\vec{0}.
\end{align}
After the second differentiation step, which is not shown for brevity, the arising terms can be simplified and their formulation can be substituted in terms of the Jacobians. This allows to factor out $\vec{u}$ and $\dot{\vec{u}}$, yielding a constraint on acceleration level\footnote{It should be observed that ${}_{B}\vec{r}_{PQ}$ is explicitly contained in these equations. In fact, the derivation yields the same result when starting with ${}_{B}\vec{r}_{PQ} = \operatorname{const.}$ In other words, the presented constraint formulation correctly captures the dynamics for arbitrary values of ${}_{B}\vec{r}_{PQ}$. This could explain why we did not experience problems due to divergence caused by numerical errors opening the kinematic loop, which, typically, is counteracted by using Baumgarte's stabilization technique~\cite{siciliano2016springer}.}
\begin{align}
\label{eq:loop_closure_constraint}
&\tilde{\mat{X}}  \vec{u} + \tilde{\mat{Y}} \dot{\vec{u}} \overset{(1,3)}{=} \vec{0}, \quad \text{where } \\
    &\tilde{\mat{X}} = \mat{R}_{BI} \big(- \skewmat{{}_{I}\vec{\omega}_{IB}} \skewmat{{}_{I}\vec{r}_{PQ}} {}_{I}\mat{J}_{IB,\mathcal{R}} + \skewmat{{}_{I}\vec{r}_{PQ}} {}_{I}\dot{\mat{J}}_{IB,\mathcal{R}} \notag \\
&\qquad - 2 \skewmat{{}_{I}\vec{\omega}_{IB}} {}_{I}\mat{J}_{PQ,\mathcal{P}} + {}_{I}\dot{\mat{J}}_{PQ,\mathcal{P}} \big),\\
 &\tilde{\mat{Y}} = \mat{R}_{BI} ( \skewmat{{}_{I}\vec{r}_{PQ}} {}_{I}\mat{J}_{IB,\mathcal{R}} + {}_{I}\mat{J}_{PQ,\mathcal{P}}).
\end{align}
Constraint \eqref{eq:loop_closure_constraint} is applied to both kinematic loops and stacked:
\begin{equation}
\label{eq:loop_closure_constraint_stacked}
    \underbrace{\begin{bmatrix}
    \tilde{\mat{X}}_l^{(1,3)} \\
    \tilde{\mat{X}}_r^{(1,3)}
    \end{bmatrix}}_{:=\mat{X}} \vec{u} + 
    \underbrace{\begin{bmatrix}
    \tilde{\mat{Y}}_l^{(1,3)} \\
    \tilde{\mat{Y}}_r^{(1,3)}
    \end{bmatrix}}_{:=\mat{Y}} \dot{\vec{u}} = \vec{0},
\end{equation}
giving the loop closure constraint in its final formulation. It will then be used in conjunction with the ground contact constraint introduced in the next section to simultaneously determine all missing constraint forces as shown in Section~\ref{subsec:equations_of_motion}.

\vspace{\subsectionspace cm} \subsection{Ground Contacts}
\label{subsec:ground_contacts}
\begin{figure}
    \label{wheel}
	\centering
	\includegraphics[width=\columnwidth]{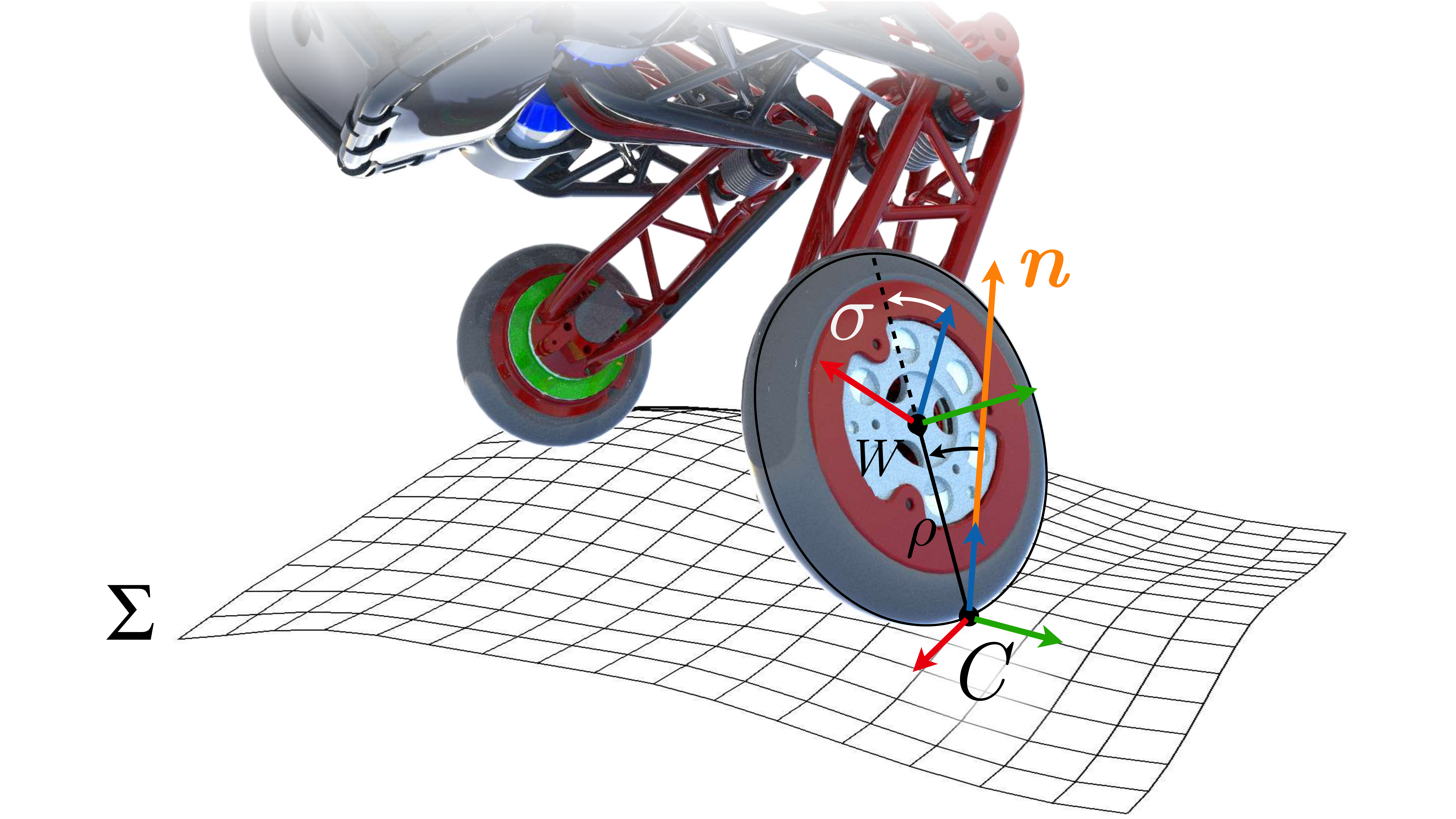}
    \vspace{-0.7cm}
	\caption{The ground contact of the (left) wheel with the ground surface $\Sigma$. $W$ denotes a wheel-fixed coordinate system and $C$ denotes the frame at the contact point on the wheel's contour, with its position parametrized by the contour parameter $\sigma$, and a fixed wheel radius $\rho$. It is oriented with its $z$-axis along the current ground normal direction $\vec{n}$ and its $x$-axis along the heading direction of the wheel.}
	\vspace{-0.5cm}
	\label{fig:wheel}
\end{figure}

To constrain the wheels' motion on the ground surface, firstly a general formulation of the rolling constraint based on rotation matrices\footnote{In contrast to e.g.~\cite{bjelonic2018keep}, where a set of local Euler angles is used.} is introduced, and then the corresponding constraint forces are derived. Again the equations are shown exemplary for one wheel and then applied to both of them.
In contrast to a point contact foot, where the area of potential contact locations is restricted to a single point, in the case of a wheel it spans the entire circumference (modeled as flat disk). Inspired by the notion of contour-kinematics introduced in~\cite{glocker}, we parametrize the position of the contact point $C$ on the contour of the wheel by a contour parameter $\sigma$, as shown in \figref{fig:wheel}. The velocity of $C$ can be expressed by
\begin{equation}
\label{eq:contact_velocity}
    {}_{I}\vec{v}_{IC}(\sigma) = {}_{I}\vec{v}_{IW} + \underbrace{\skewmat{{}_{I}\vec{\omega}_{IW}} {}_{I}\vec{r}_{WC}(\sigma)}_{{}_{I}\vec{v}_{WC}(\sigma)},
\end{equation}
\begin{equation} 
\label{eq:relative_contact_position}
    \text{where} \quad {}_{I}\vec{r}_{WC}(\sigma) = \mat{R}_{IW} \underbrace{
    \begin{bmatrix}
    \rho \sin(\sigma) & 0 & \rho \cos(\sigma)
    \end{bmatrix}^\top
    }_{{}_{W}\vec{r}_{WC}(\sigma)},
\end{equation}
where $\rho$ denotes the wheel radius, and where ${}_{I}\vec{v}_{IW}$, ${}_{I}\vec{\omega}_{IW}$ and $\mat{R}_{IW}$ can be obtained from forward differential kinematics. The contour parameter $\sigma$ can directly be calculated from the current normal vector of the ground surface:
\begin{equation}
\label{eq:sigma}
    \sigma({}_{I}\vec{n}) = \operatorname{arctan2}(\mat{R}_{WI}^{(1)} \, {}_{I}\vec{n}, \; \mat{R}_{WI}^{(3)} \, {}_{I}\vec{n}).
\end{equation}
To impose a constraint on the velocity of $C$ that can be used to find the contact forces, the constraint must be formulated on the acceleration level. We therefore differentiate \eqref{eq:contact_velocity} w.r.t. time, which gives
\begin{equation}
\label{eq:contact_acceleration}
    {}_{I}\vec{a}_{IC}(\sigma, \; \dot{\sigma}) = {}_{I}\vec{a}_{IW} + 
    {}_{I}\vec{a}_{WC}(\sigma, \; \dot{\sigma}),
\end{equation}
where from ${}_{I}\vec{v}_{WC}$, as given in \eqref{eq:contact_velocity}, we obtain\footnote{We note that ${}_{W}\vec{t}$ points along the current tangential direction of the wheel contour. The expression containing ${}_{W}\vec{t}$ can be interpreted as a compensation term for the centripetal acceleration that a wheel-fixed point experiences as it moves with the wheel.}
\begin{multline}
    {}_{I}\vec{a}_{WC}(\sigma, \; \dot{\sigma}) 
    = \skewmat{{}_{I}\dot{\vec{\omega}}_{IW}} {}_{I}\vec{r}_{WC}(\sigma)
    + \skewmat{{}_{I}\vec{\omega}_{IW}}^2 {}_{I}\vec{r}_{WC}(\sigma) \\
    + \skewmat{{}_{I}\vec{\omega}_{IW}} \mat{R}_{IW} \underbrace{\frac{d}{d\sigma}{}_{W}\vec{r}_{WC}(\sigma)}_{:={}_{W}\vec{t}} \dot{\sigma}.
\end{multline}
A closed form expression for $\dot{\sigma}$ can be found by differentiating \eqref{eq:sigma}, which is made possible by the continuous differentiability of $\operatorname{arctan2}$. Application of the chain rule gives
\begin{equation}
\label{eq:sigma_dot}
    \dot{\sigma}({}_{I}\vec{n}, \; {}_{I}\vec{\dot{\vec{n}}})
    = \frac{d\sigma}{d(\mat{R}_{WI} \, {}_{I}\vec{n})}
    (\skewmat{{}_{W}\vec{\omega}{}_{WI}} \mat{R}_{WI} \, {}_{I}\vec{n}
    +  \mat{R}_{WI} \, {}_{I}\dot{\vec{n}} ).
\end{equation}
If ${}_{I}\dot{\vec{n}}$ is assumed to be $\vec{0}$,\footnote{If this is not the case, ${}_{I}\dot{\vec{n}}$ can be appended to the generalized velocity vector $\vec{u}$ to achieve the linear Jacobian relationship of \eqref{eq:contact_acceleration_Jacobians}. We note that the current ground surface estimation can then be formulated as a constrained state estimation problem. However, this is beyond the scope of this letter.} $\vec{u}$ and $\dot{\vec{u}}$ can be factored out of \eqref{eq:contact_acceleration}, resulting in the Jacobian formulation
\begin{equation}
\label{eq:contact_acceleration_Jacobians}
    {}_{I}\vec{a}_{IC}(\sigma, \; \dot{\sigma}) 
    = {}_{I}\dot{\mat{J}}_{IC,\mathcal{P}}(\sigma) \, \vec{u}
    + {}_{I}\mat{J}_{IC,\mathcal{P}}(\sigma) \, \dot{\vec{u}}.
\end{equation}
For perfect rolling, a zero acceleration constraint must be enforced in the $x$- and $z$-directions of the contact frame, i.e.
\begin{equation}
\label{eq:rolling_constraint}
    {}_{C}\dot{\mat{J}}_{IC,\mathcal{P}}(\sigma) \, \vec{u}
    + {}_{C}\mat{J}_{IC,\mathcal{P}}(\sigma) \, \dot{\vec{u}} \overset{(1,3)}{=} \vec{0}.
\end{equation}
All of the above derivations can be performed analogously for the left- and the right-hand side, which allows stacking of the obtained quantities:
\begin{align}
\label{eq:rolling_constraint_stacked}
    \underbrace{\begin{bmatrix}
     {}_{C_l}\mat{J}_{IC_l,\mathcal{P}}^{(1,3)} \\
     {}_{C_r}\mat{J}_{IC_r,\mathcal{P}}^{(1,3)}
    \end{bmatrix}}_{:=\mat{J}_{A}} \vec{u} + 
    \underbrace{\begin{bmatrix}
    {}_{C_l}\dot{\mat{J}}_{IC_l,\mathcal{P}}^{(1,3)} \\
    {}_{C_r}\dot{\mat{J}}_{IC_r,\mathcal{P}}^{(1,3)}
    \end{bmatrix}}_{:=\dot{\mat{J}}_{A}} \dot{\vec{u}} = \vec{0}.
\end{align}{}

As the robot is capable of leaning to its sides, the distance of the two contact points of the wheels does not remain constant. Therefore, slipping occurs in the $y$-directions of the contact frames, leading to friction forces acting along 
\begin{equation}
\mat{J}_F := \begin{bmatrix} {}_{C_l}\mat{J}^{(2)}_{IC_l,\mathcal{P}} \\
{}_{C_r}\mat{J}^{(2)}_{IC_r,\mathcal{P}} \end{bmatrix}.
\end{equation}
Now, rolling constraint forces $\vec{F}_{C} \in \mathbb{R}^{4}$ and friction terms are added to \eqref{eq:loop_closure_dynamics_compressed}:
\begin{equation}
\label{eq:full_dynamics}
\mat{M} \, \dot{\vec{u}} + \vec{b} + \vec{g} + \vec{s} + \mat{J}_{L}^\top \, \vec{F}_{L} + \mat{J}_{A}^\top \, \mat{F}_C + \mat{J}_{F}^\top \, \mat{C}_{F} \, \vec{F}_{C}  = \mat{S}^\top \vec{\tau},
\end{equation}
where $\mat{C}_F$ represents a velocity dependent friction curve which we model by the differentiable $\operatorname{tanh}$-function, with $\mu_{s}$ being the sliding friction coefficient between tires and ground:
\begin{align}
    \mat{C}_F = -\mu_{s} \begin{bmatrix}
    0 & \tanh(\mat{J}_F^{(1)} \, \vec{u}) & 0 & 0 \\ 0 & 0 & 0 & \tanh(\mat{J}_F^{(2)} \, \vec{u}) 
    \end{bmatrix}.
\end{align}

\vspace{\subsectionspace cm} \subsection{Solving for the Unknown Constraint Forces}
\label{subsec:equations_of_motion}
To state the complete \ac{EoM}, all that is left is to determine the unknown constraint forces in \eqref{eq:full_dynamics} with the aid of the constraints \eqref{eq:loop_closure_constraint_stacked} and \eqref{eq:rolling_constraint_stacked}. For this purpose, we first stack the unknown forces and the corresponding constraints:
\begin{align}
    \mat{M} \dot{\vec{u}} + \vec{b} + \vec{g} + \vec{s} + 
    {\underbrace{\begin{bmatrix}
    \mat{J}_{L} \\
    \mat{J}_{A} + \mat{C}_F^\top \mat{J}_{F} 
    \end{bmatrix}}_{:=\mat{J}}}^\top
    \underbrace{\begin{bmatrix}
    \vec{F}_L \\
    \vec{F}_C
    \end{bmatrix}}_{:=\vec{F}}
    &=\mat{S}^{\top} \vec{\tau},\label{eq:full_dynamics_compressed}\\
    \underbrace{\begin{bmatrix}
    \mat{X}^\top &
    \dot{\mat{J}}_{A}^\top
    \end{bmatrix}^\top}_{:=\mat{V}} \vec{u} +
    \underbrace{\begin{bmatrix}
    \mat{Y}^\top &
    \mat{J}_{A}^\top
    \end{bmatrix}^\top}_{:=\mat{W}}  \dot{\vec{u}} &= \vec{0}.\label{eq:constraints_compressed}
\end{align}
Next, \eqref{eq:full_dynamics_compressed} can be solved for $\dot{\vec{u}}$, which can be inserted in \eqref{eq:constraints_compressed} and then solved for the constraint forces, resulting in
\begin{equation}
\label{eq:missing_forces}
        \mat{F} = (\mat{W} \mat{M}^{-1} \mat{J}^{\top})^{-1} (\mat{V} \vec{u} + \mat{W} \mat{M}^{-1} (\mat{S}^{\top} \vec{\tau} - \vec{b} - \vec{g} - \vec{s})).
\end{equation}
By substituting, the final \ac{EoM} of the system are obtained.

\vspace{\sectionspace cm} \section{Control}
\label{sec:control}
    \vspace{\subsectionspace cm} \subsection{Whole-Body Control}
\label{subsec:whole-body-control}
The \ac{WBC}-problem in its basic form can be written as a hierarchical quadratic optimization, see for instance~\cite{bellicos2016perception},
\begin{align}
\label{eq:wbc}
    \vec{x}_{i} = \underset{\vec{x}}{\operatorname{argmin}} \ &\| \mat{A}_i \, \vec{x} - \vec{b}_i\|_{2}^{2} \\
\label{eq:wbc_constraints}
    \operatorname{s. t.} 
    \begin{bmatrix}
    \mat{A}_1 \\
    \vdots \\
    \mat{A}_{i-1}
    \end{bmatrix} \, \vec{x} = 
    \begin{bmatrix}
    \vec{A}_1 \, \vec{x}_1 \\
    \vdots \\
    \vec{A}_{i-1} \, \vec{x}_{i-1}
    \end{bmatrix}, &
    \begin{bmatrix}
    \mat{C}_1 \\
    \vdots \\
    \mat{C}_{n_{ineq}}
    \end{bmatrix} \, \vec{x} \leq 
    \begin{bmatrix}
    \vec{d}_1 \\
    \vdots \\
    \vec{d}_{n_{ineq}}
    \end{bmatrix},
\end{align}
where in each iteration $i$ a task is added. In the context of this work, a task is defined as a linear equality $\mat{A}_i \, \vec{x} = \vec{b}_i$ which is to be satisfied as accurately as possible with regard to the $2$-norm. By expanding the objective, each iteration of \eqref{eq:wbc} subject to \eqref{eq:wbc_constraints} is formulated as a \ac{QP} 
\begin{align}
    \underset{\vec{x}}{\operatorname{argmin}} \ \frac{1}{2} \vec{x}^\top \mat{H} \vec{x} + \vec{f}^{\top}\vec{x} \quad  \operatorname{s. t.} \; \mat{A} \vec{x} = \vec{b}, \; \mat{C}  \vec{x} \leq \vec{d},
\end{align}
where we define the optimization variables as
\begin{equation}
    \vec{x} =
    \begin{bmatrix}
    \dot{\vec{u}}^{\top} & \vec{F}_L^{\top} & \vec{F}_C^{\top} & \vec{\tau}^{\top}
    \end{bmatrix}^{\top}.
\end{equation}
After completing the last iteration, the actuation torques $\vec{\tau}$ are directly applied to the system.

\vspace{\subsectionspace cm} \subsection{Motion, Force and Torque Tasks}
In the following, we list the motion, force, and torque tasks in hierarchical order, from highest to lowest priority.

\subsubsection{Dynamics Model}
\label{subsec:dynamic_model}
In order for the motion to remain physical, the optimization variables in $\vec{x}$ must satisfy the constrained \ac{EoM} \eqref{eq:full_dynamics_compressed} and \eqref{eq:constraints_compressed}:
\begin{align}
    \mat{A}_1 = 
    \begin{bmatrix}
    \mat{M} & \mat{J}_L^\top & \mat{J}_A^\top + \mat{J}_F^\top \, \mat{C}_F  & -\mat{S}^\top \\
    \mat{W} & \mat{0} & \mat{0} & \mat{0}
    \end{bmatrix}, & \
    \vec{b}_1 = 
    \begin{bmatrix}
    -\vec{b}-\vec{g}-\vec{s}\\
    -\mat{V} \vec{u}
    \end{bmatrix}.
\end{align}

\subsubsection{Base Height}
\label{subsubsec:base_height}
This task controls the height of the base w.r.t. a local control frame $N$ -- similar to the one in~\cite{bellicoso2017dynamic} -- which has $x$-axis aligned with the robot's heading direction, $y$-axis pointing along the \ac{LoS}, and origin at the midpoint $G$ thereof. The task on acceleration level is given by
\begin{align}
    \mat{A}_2 = 
    \begin{bmatrix}
    {}_{N}\mat{J}_{NB,\mathcal{P}}^{(3)} & \mat{0} & \mat{0} & \mat{0}
    \end{bmatrix}, & \
    \vec{b}_2 = 
    {}_{\ N}^{des}\ddot{\vec{r}}_{NB}^{(3)} - {}_{N}\dot{\mat{J}}_{NB,\mathcal{P}}^{(3)} \vec{u},
\end{align}
where the desired operational space acceleration ${}_{\ N}^{des}\ddot{\vec{r}}_{NB}^{(3)}$ is controlled along a reference trajectory by a \ac{PD}-law
\begin{multline}
\label{eq:pd_law}
    {}_{\ N}^{des}\ddot{\vec{r}}_{NB}^{(3)} =
    k_{p}({}_{\ N}^{ref}\vec{r}_{NB}^{(3)} - {}_{N}\vec{r}_{NB}^{(3)}) \\
    + k_{d}({}_{\ N}^{ref}\dot{\vec{r}}_{NB}^{(3)}  - {}_{N}\dot{\vec{r}}_{NB}^{(3)})
    + {}_{\ N}^{ref}\ddot{\vec{r}}_{NB}^{(3)},
\end{multline}
where $k_p$ and $k_d$ denote the respective gains. We modify ${}_{\ N}^{ref}\vec{r}_{NB}^{(3)}$ to take the current roll and pitch angles into account, in order to let the system behave like an inverted pendulum with fixed length.

\subsubsection{Base Roll Angle}
\label{subsubsec:base_roll}
\begin{figure}
	\centering
	\includegraphics[width=\columnwidth]{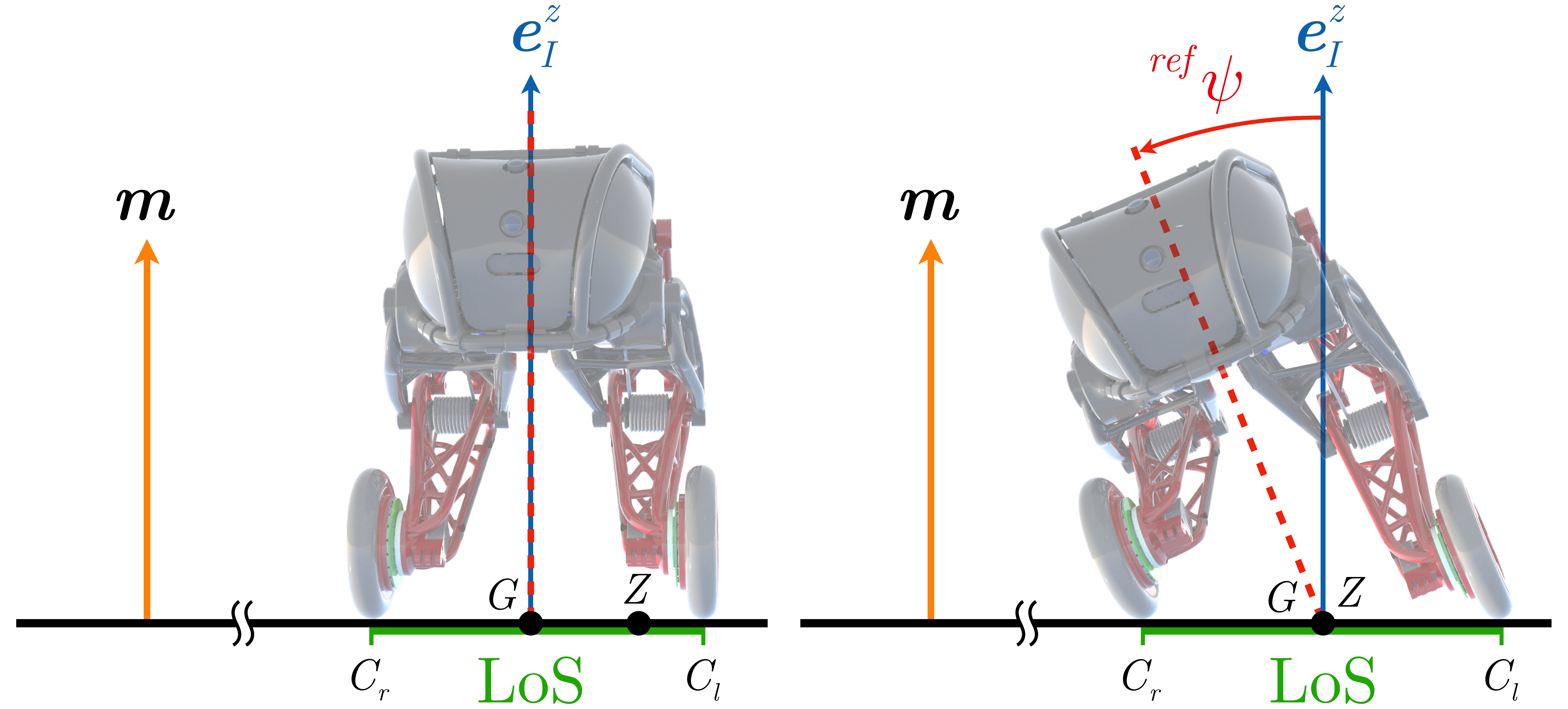}
    \vspace{-0.7cm}
	\caption{Curve driving of the robot around the axis of rotation $\vec{m}$. \emph{Left:} The system drives without adjusting its roll-angle reference ${}^{ref}\psi$. The larger the linear and angular velocities are, the more the \ac{ZMP} moves outwards towards the endpoint of the current \ac{LoS}, causing the robot to gradually lose robustness against tipping over. \emph{Right:} The roll-angle reference ${}^{ref}\psi$ is adjusted such that the \ac{ZMP} coincides with the center $G$ of the \ac{LoS}, thereby increasing the robustness of the robot against falling sideways.}
	\vspace{-0.5cm}
	\label{fig:zmp}
\end{figure}

This task controls the roll angle $\psi$ of the base. This allows the robot to maintain a prespecified roll orientation, even if the extensions of the legs change, for instance due to unmodeled uneven terrain, as shown in \ref{subsubsec:adaptive_legs}. The task on acceleration level is given by
\begin{align}
    \mat{A}_3 = 
    \begin{bmatrix}
    {}_{N}\mat{J}_{NB,\mathcal{R}}^{(1)} & \mat{0} & \mat{0} & \mat{0}
    \end{bmatrix}, & \
    \vec{b}_3 = 
    {}^{des}\ddot{\psi} - {}_{N}\dot{\mat{J}}_{NB,\mathcal{R}}^{(1)} \vec{u},
\end{align}
where the desired roll acceleration ${}^{des}\ddot{\psi}$ is again controlled along a reference trajectory by a \ac{PD}-law analogous to \eqref{eq:pd_law}.
For a bipedal system, driving tight curves is possible, but can lead to loss of robustness against tipping over. To counteract this effect we compute the roll angle reference such that the \ac{ZMP} comes to lie at $G$,\footnote{The centering of the \ac{ZMP} could also be achieved by adding a task requiring the normal forces of both wheels to be equal, but the chosen approach enables tuning of the leaning aggressiveness. Further, it would also be possible to tune the equality constraint on the normal forces by adding it to the objective of the \ac{WBC} as a soft constraint.} as shown in \figref{fig:zmp}.

\subsubsection{LQR-Assisted Balancing}
\begin{figure}
	\centering
	\includegraphics[width=\columnwidth]{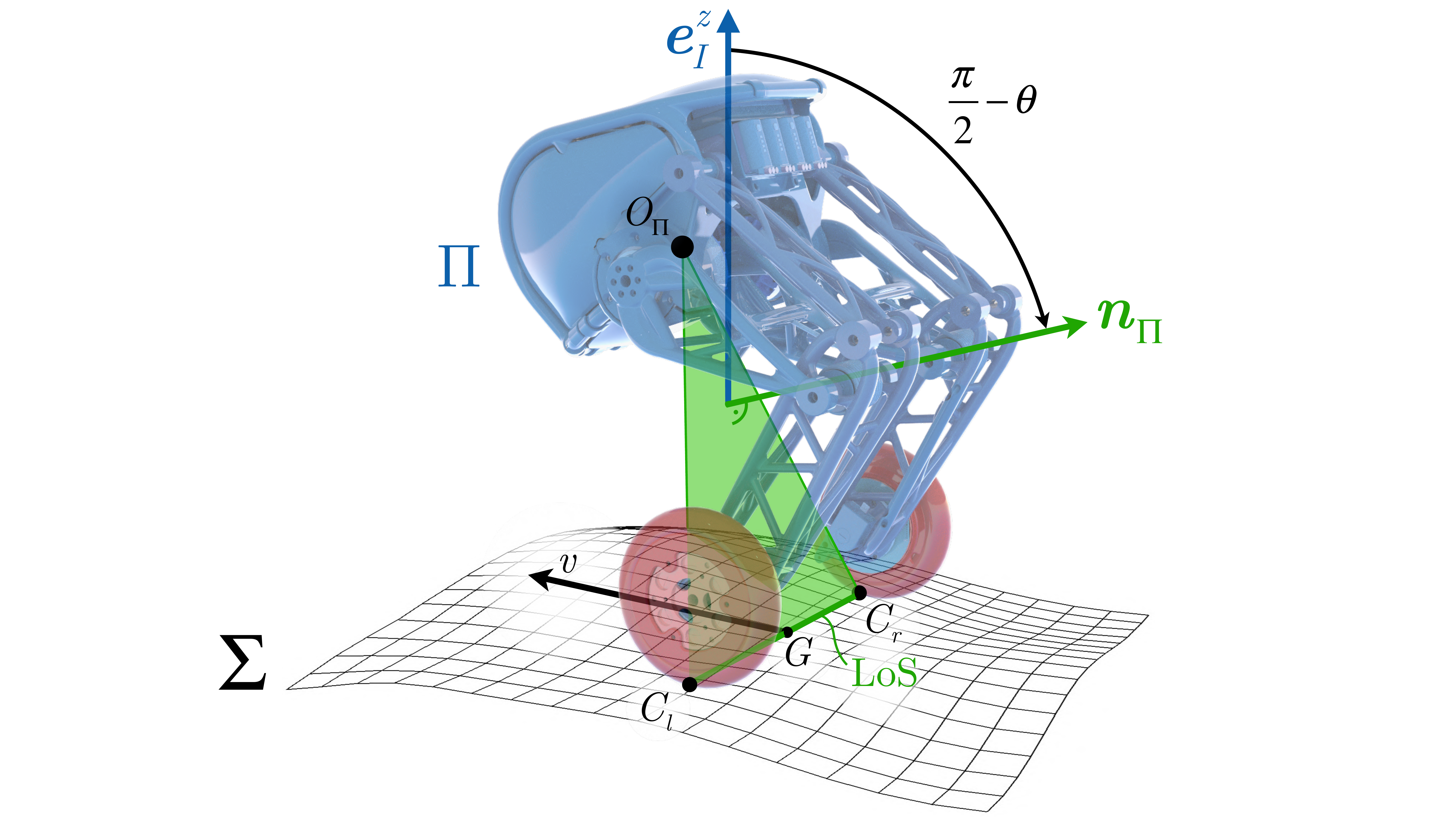}
    \vspace{-0.7cm}
	\caption{Relevant quantities of the simplified system used for synthesis of the \ac{LQR} control law. The base and the leg structures (i.e. all rigid bodies except the wheels) are lumped into a substitute pendulum body $\mathit{\Pi}$. The pitch angle $\theta$ of $\mathit{\Pi}$ can be found by spanning a plane with normal $\vec{n}_{\mathit{\Pi}}$ between the \ac{CoM} of $\mathit{\Pi}$ ($O_{\mathit{\Pi}}$) and the two contact points of the wheels. Thereof, $\theta$ can be extracted such that $\theta = 0$ if $\vec{n}_{\mathit{\Pi}}$ and $\vec{e}_{I}^{z}$ (i.e. the $z$-direction of the $I$ frame) are orthogonal, or in other words, if $O_{\mathit{\Pi}}$ lies above the \ac{LoS}. This notion allows stability also when balancing with wheels on different heights (shown in Fig.~\ref{fig:exp_leg}), or when leaning into curves (as in Fig.~\ref{fig:exp_lean}). Further, $v$ is defined as the average wheel hub velocity in the current heading direction and can be interpreted as the ground velocity of the system.}
	\vspace{-0.5cm}
	\label{fig:lqr}
\end{figure}

The direct application of \ac{WBC} to systems with non-minimum phase dynamics, as in the case of a wheeled balancing robot, can be problematic. This is illustrated by the following example: For the base to accelerate forwards from the upright position, the wheels first need to accelerate backwards to achieve sufficient pitch angle and should only then accelerate forwards, to avoid falling over. In its standard form, \ac{WBC} fails to reproduce this behavior since it computes accelerations in the direction of the desired motion only. We propose to overcome this issue by including an \ac{LQR} feedback law as a motion task, shown in the following. 
Firstly, a simplified model (seen in \figref{fig:lqr}) is created which captures the essential dynamics of a two-wheeled inverted pendulum system, i.e. the coupling of the tilting and driving motions. This results in a lumped pendulum body $\mathit{\Pi}$ with pitch angle $\theta$ and average wheel hub velocity $v$. We define ${}^{des}\ddot{\theta}$ as the input to the simplified system, as this is the quantity the \ac{WBC} will be tracking. The simplified system is then linearized at the current operating point,\footnote{The required accelerations $\dot{\vec{u}}$ are calculated using \eqref{eq:full_dynamics_compressed}, \eqref{eq:missing_forces} from the current system state $\vec{q}$, $\vec{u}$.} resulting in a state-space system of the form
\begin{equation}
\label{eq:simplified_state_space_model}
 \begin{bmatrix}
        \dot{\theta} \\ \ddot{\theta} \\ \dot{v}
    \end{bmatrix} = 
        \begin{bmatrix}
        0 & 1 & 0 \\
        0 & 0 & 0 \\
        a_{3,1} & a_{3,2} &  0
    \end{bmatrix}
        \begin{bmatrix}
 \theta \\ \dot{\theta} \\ v
    \end{bmatrix} + 
        \begin{bmatrix}
        0 \\ 1 \\ b_{3,1}
    \end{bmatrix}
    {}^{des}\ddot{\theta},
\end{equation}
where $a_{3,1}$, $a_{3,2}$, and $b_{3,1}$ are functions of the lumped inertias, lumped masses and the lumped pendulum length, evaluated at each time step $T_s$ of the controller. \eqref{eq:simplified_state_space_model} is then discretized, assuming \ac{ZOH} over $T_s$. Finally, the \ac{DARE} is solved for the discretized system, yielding the infinite-horizon controller gain matrix $ \mat{K} = \begin{bmatrix} k_{\theta} & k_{\dot{\theta}} & k_{v} \end{bmatrix}$.
The corresponding feedback law
\begin{equation}
\label{eq:lqr_feedback_law}
    {}^{des}\ddot{\theta} = k_{\theta}({}^{ref}\theta - \theta) + k_{\dot{\theta}}({}^{ref}\dot{\theta} - \dot{\theta}) + k_{v}({}^{ref}v - v)
\end{equation}
can be used to track a reference trajectory, e.g. a positive ground velocity. The controller will then respond with the desired motion, i.e. first driving backwards a little and then accelerating forwards.

To include \eqref{eq:lqr_feedback_law} in a \ac{WBC}-task, we track ${}^{des}\ddot{\theta}$ by deriving an approximated lumped rotational Jacobian ${}_{N}\mat{J}_{N\mathit{\Pi},\mathcal{R}}$ for the lumped pendulum body $\mathit{\Pi}$. We start by using the notion of average angular momentum introduced in~\cite{essen1993average} to lump all bodies ($n_{bod}$) except for the two wheels together, approximating it as
\begin{equation}
\label{eq:average_angular_momentum}
    {}_{I}\mat{I}_{\mathit{\Pi}} \, {}_{I}\vec{\omega}_{I\mathit{\Pi}} =
    \sum_{K=1}^{n_{bod}-2} {}_{I}\mat{I}_{K} \, {}_{I}\vec{\omega}_{I\mathit{\Pi}} =
    \sum_{K=1}^{n_{bod}-2} {}_{I}\mat{I}_{K} \, {}_{I}\vec{\omega}_{IK},
\end{equation}
where ${}_{I}\mat{I}_{K}$ denotes the inertia tensor at the \ac{CoM}-frame of body $K$ represented in frame $I$. By rewriting \eqref{eq:average_angular_momentum} in Jacobian form, the lumped rotational Jacobian can be defined as
\begin{equation}
    {}_{I}\mat{J}_{I\mathit{\Pi},\mathcal{R}} = ({}_{I}\mat{I}_{\mathit{\Pi}})^{-1}  \sum_{K=1}^{n_{bod}-2} {}_{I}\mat{I}_{K} \, {}_{I}\mat{J}_{IK,\mathcal{R}}.
\end{equation}
Similarly, ${}_{I}\dot{\mat{J}}_{I\mathit{\Pi},\mathcal{R}}$ is found, which enables the formulation of a motion task for $\ddot{\theta}$ in the control frame $N$ as
\begin{align}
    \mat{A}_4 = 
    \begin{bmatrix}
    {}_{N}\mat{J}_{N\mathit{\Pi},\mathcal{R}}^{(2)} & \mat{0} & \mat{0} & \mat{0}
    \end{bmatrix}, & \
    \vec{b}_4 = 
    {}^{des}\ddot{\theta} - {}_{N}\dot{\mat{J}}_{N\mathit{\Pi},\mathcal{R}}^{(2)} \vec{u}.
\end{align}

\subsubsection{Base Yaw Angle}
\label{subsubsec:base_yaw}
This task controls the yaw angle $\phi$ of the base, and therefore the robot's heading direction. The task on acceleration level is given by
\begin{align}
    \mat{A}_5 = 
    \begin{bmatrix}
    {}_{N}\mat{J}_{NB,\mathcal{R}}^{(3)} & \mat{0} & \mat{0} & \mat{0}
    \end{bmatrix}, & \
    \vec{b}_5 = 
    {}^{des}\ddot{\phi} - {}_{N}\dot{\mat{J}}_{NB,\mathcal{R}}^{(3)} \vec{u},
\end{align}
whereby the desired yaw acceleration ${}^{des}\ddot{\phi}$ is controlled by a feedback law analogous to \eqref{eq:pd_law}.

\subsubsection{Actuation Torque Minimization}
\label{subsubsec:actuation_torque_minimization}
A unique solution is enforced by minimizing all actuation torques (and thereby also the robot's power consumption):
\begin{align}
    \mat{A}_6 = 
    \begin{bmatrix}
    \mat{0} & \mat{0} & \mat{0} & \mathbb{I}_{n_{\tau} \times n_{\tau}}
    \end{bmatrix}, & \
    \vec{b}_6 = \vec{0},
\end{align}
where $\mathbb{I}_{n_{\tau} \times n_{\tau}}$ denotes the $n_{\tau} \times n_{\tau}$ identity matrix.

\vspace{\subsectionspace cm} \subsection{Inequality Constraints}
\label{subsec:inequality_constraints}
In the following, we list the inequality constraints contributing to \eqref{eq:wbc_constraints}. Since they are enforced at every hierarchy level, their ordering does not matter.

\subsubsection{Actuator Saturation}
\label{subsubsec:actuator_saturation}
Joint torques are constrained by the bi-directional actuator saturation bounds $-{}^{sat}\vec{\tau} \leq \vec{\tau} \leq {}^{sat}\vec{\tau}$.

\subsubsection{Unilateral Contact Forces}
\label{subsubsec:no_lift_off}
To prevent the robot from \enquote{pulling} on the ground, we add the following two inequality constraints $\vec{F}_{C}^{(2)}\leq 0$ and $\vec{F}_{C}^{(4)}\leq 0$.

\subsubsection{Static Friction}
\label{subsubsec:static_friction}
To prevent slipping in rolling direction, we constrain the corresponding friction forces to stay within the static friction bounds $\pm\vec{F}_{C}^{(1)}\leq -\mu_{h} \vec{F}_{C}^{(2)}$ and $\pm\vec{F}_{C}^{(3)}\leq -\mu_{h} \vec{F}_{C}^{(4)}$,
where $\mu_{h}$ is the coefficient of static friction.

\vspace{\sectionspace cm} \section{Results and Discussion}
\label{sec:results_and_discussion}
    The modeling and control approaches presented in this work were first validated in a custom \emph{MATLAB} simulation and in \emph{Gazebo}~\cite{gazebo} with \emph{ODE}~\cite{ode} as physics back-end, and then tested on hardware, that is, the \emph{Ascento} robot. In the following, we outline the implementation details of the control pipeline -- including the state estimation -- and show three selected experiments. These, and additional ones, are also presented in the accompanying video.

\vspace{\subsectionspace cm} \subsection{Setup}
\label{subsec:setup}

\subsubsection{State Estimation}
Through the robot's sensors, i.e. an \ac{IMU} (with an integrated filter) mounted to the base and rotational encoders at each of the four motors, we can kinematically reconstruct the state of the robot up to its absolute position $\vec{r}_{IB}$ and velocity $\vec{v}_{IB}$.\footnote{Estimation of the missing quantities by integration leads to poor results due to the high measurement noise of the \ac{IMU}.} We therefore use the contact point of the left wheel as a reference, inspired by~\cite{bloesch2012state}, getting its absolute position and velocity through a wheel odometry estimate, assuming ground contact for all times. To reflect this choice also in the model, we adjust the contact situation at the wheels by removing sliding friction and adding a (slightly unphysical) hard contact constraint in the $y$-direction of the left wheel's contact frame. Approaches to circumvent this problem in the scope of future work are discussed in Section~\ref{sec:conclusion_and_outlook}. Further, the ground is considered locally flat for modeling.
\begin{figure}[!t]
	\centering
	\includegraphics[width=\columnwidth]{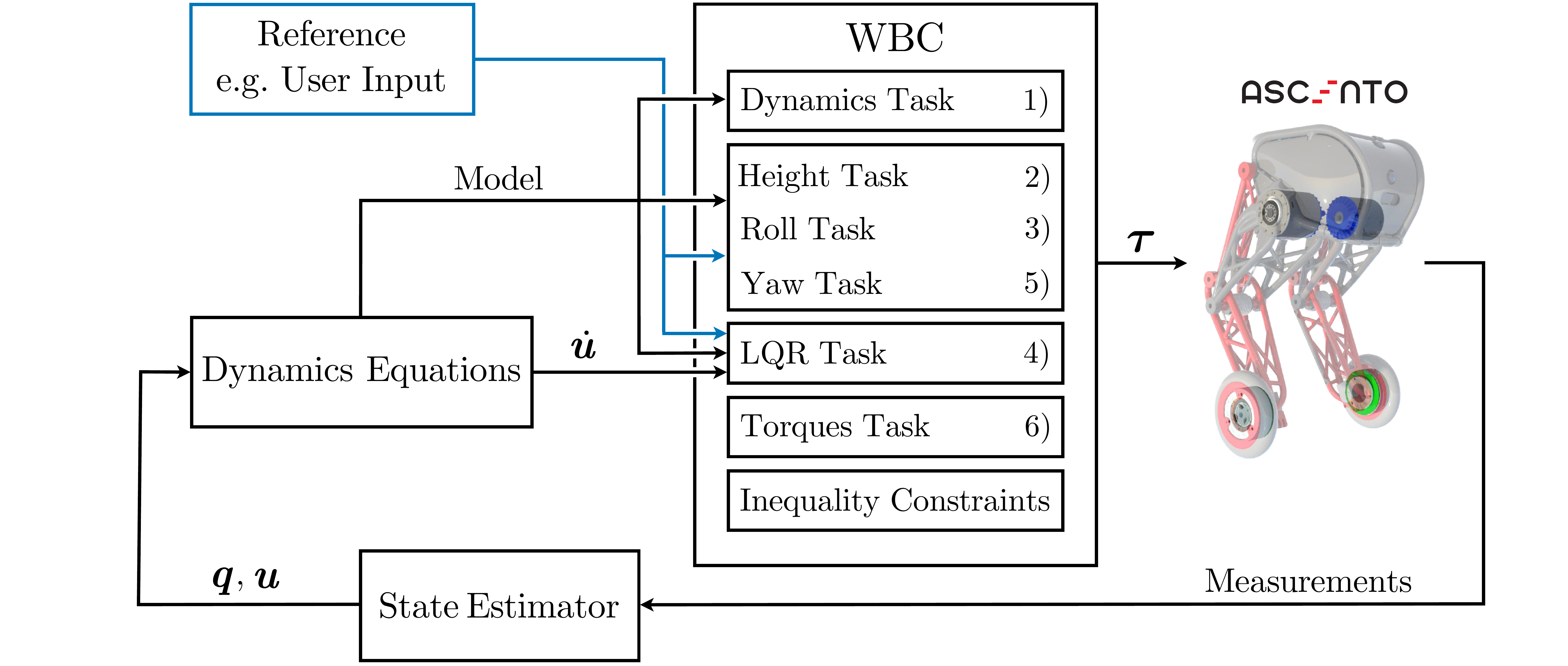}
    \vspace{-0.7cm}
	\caption{Block diagram of the control pipeline running on the robot.} 
	\vspace{-0.5cm}
	\label{fig:diagram}
\end{figure}

\subsubsection{Implementation}
\label{implementation}
The dynamics model\footnote{The geometrical and inertial parameters were obtained from the \ac{CAD} model of \emph{Ascento}. Further, a static friction coefficient $\mu_h$ of \SI{0.8}{} has been assumed, which is typical for tire on road conditions.} and the \ac{WBC}\footnote{The \ac{PD} gains for tasks 2), 3) and 5) were tuned each by first adjusting the proportional gain $k_{p}$ to follow a trajectory with desired aggressiveness. Next, the derivative gain was selected as $k_{d} = 2 \sqrt{k_{p}}$, which corresponds to ideal damping for a unitary mass harmonic oscillator. In the cases where this choice of $k_p$ lead to oscillations caused by noisy state estimates, we reduced $k_p$ until these would disappear. The cost matrices for state and input cost of the \ac{LQR} in task 4) were tuned similarly as presented in \cite{ascento_paper} since the simplified state-space model \eqref{eq:simplified_state_space_model} is a subset of the full model in \cite{ascento_paper}.} were implemented in ROS/C++ using \emph{Eigen} as linear algebra library~\cite{eigen}. For the dynamics model -- in particular the computation of the Jacobians -- we used a custom formulation that exploits recursive dependencies and maximizes the reuse of quantities that appear multiple times. The \acp{QP} arising from the \ac{WBC} scheme are solved using the state of the art \ac{QP}-solver \emph{OSQP}~\cite{stellato2017osqp}. For the solution of the \ac{DARE} for the \ac{LQR}, the \emph{Control Toolbox}~\cite{adrlCT} library was employed. The \ac{QP} and the \ac{DARE} get automatically warm-started with the solution of the previous time step. We run the controller at a frequency of \SI{400}{Hz} on the onboard computer.\footnote{I.e. a NUC7I7BNH with \SI{3.5}{GHz} dual-core Intel Core i7 processor.} The average control period is \SI{1.56}{ms}, whereof \SI{1.20}{ms} are used for the hierarchical optimization of the \ac{WBC}, \SI{0.11}{ms} are used for the evaluation of the model, and the remaining time is attributed to the calculation of the \ac{DARE}, the state estimator and program overhead. An illustration of the control loop is shown in \figref{fig:diagram}.

\vspace{\subsectionspace cm} \subsection{Experiments}
\label{subsec:experiments}

\subsubsection{Impact Robustness of Balancing Control}
\label{subsubsec:balancing_performance}
\begin{figure}
	\centering
	\vspace{-0.3cm}
	\includegraphics[width=\columnwidth]{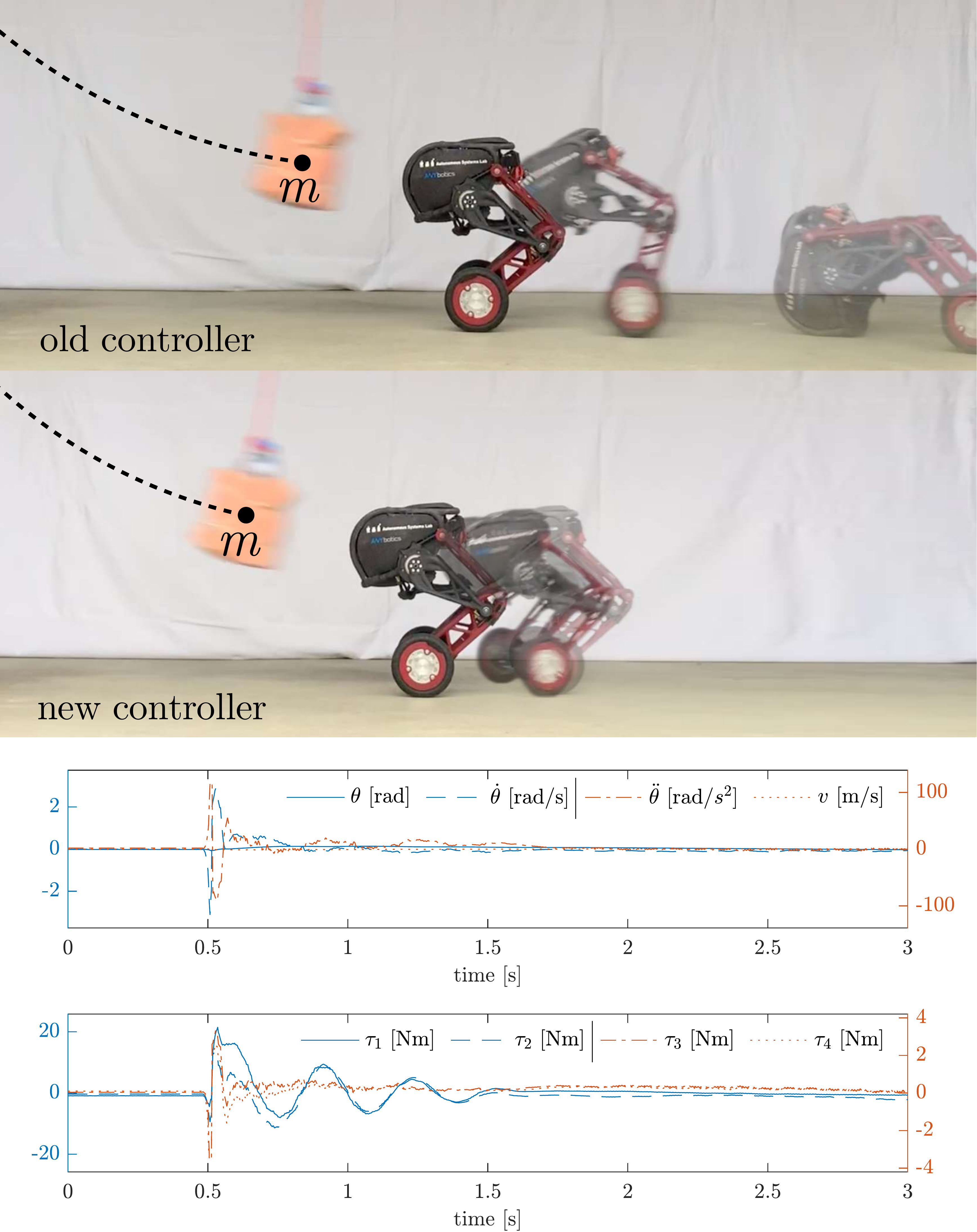}
    \vspace{-0.7cm}
	\caption{Impact robustness of balancing control. The images on the top qualitatively show the response to an impact with a $m=\SI{2}{kg}$ load for the previous \ac{LQR}-based controller~\cite{ascento_paper} and for the \ac{WBC} scheme proposed in this work. The two graphs below quantitatively show the corresponding response for the proposed \ac{WBC} scheme. The actuation torques (as introduced in \figref{fig:gen_cor}) computed by the \ac{WBC} are depicted in the bottom graph and the resulting evolution of the ground velocity $v$ and the base pitch angle $\theta$ in the upper graph. It is to be noted that the angular acceleration response $\ddot{\theta}$ was calculated based on the current system state $\vec{q}$, $\vec{u}$ using \eqref{eq:full_dynamics_compressed}, \eqref{eq:missing_forces}.}
	\vspace{-0.5cm}
    \label{fig:exp_bal}
\end{figure}

A \SI{2}{kg} weight attached to a cord is dropped from a relative height of \SI{1}{m} to create a horizontal impact with the robot, as shown in \figref{fig:exp_bal}. The response is qualitatively compared to the previous, \ac{LQR}-based controller\footnote{This controller assumed a two-wheeled, inverted pendulum of constant length as model and used a fixed linearization point around the upright equilibrium.} running on \emph{Ascento}~\cite{ascento_paper}, which fails to stabilize the system, while the controller proposed in this work recovers from the disturbance with a T90 time of ca.~\SI{1}{s}.

\subsubsection{Adaption to Varying Ground Heights}
\label{subsubsec:adaptive_legs}
\begin{figure}
	\centering
	\vspace{-0.3cm}
	\includegraphics[width=\columnwidth]{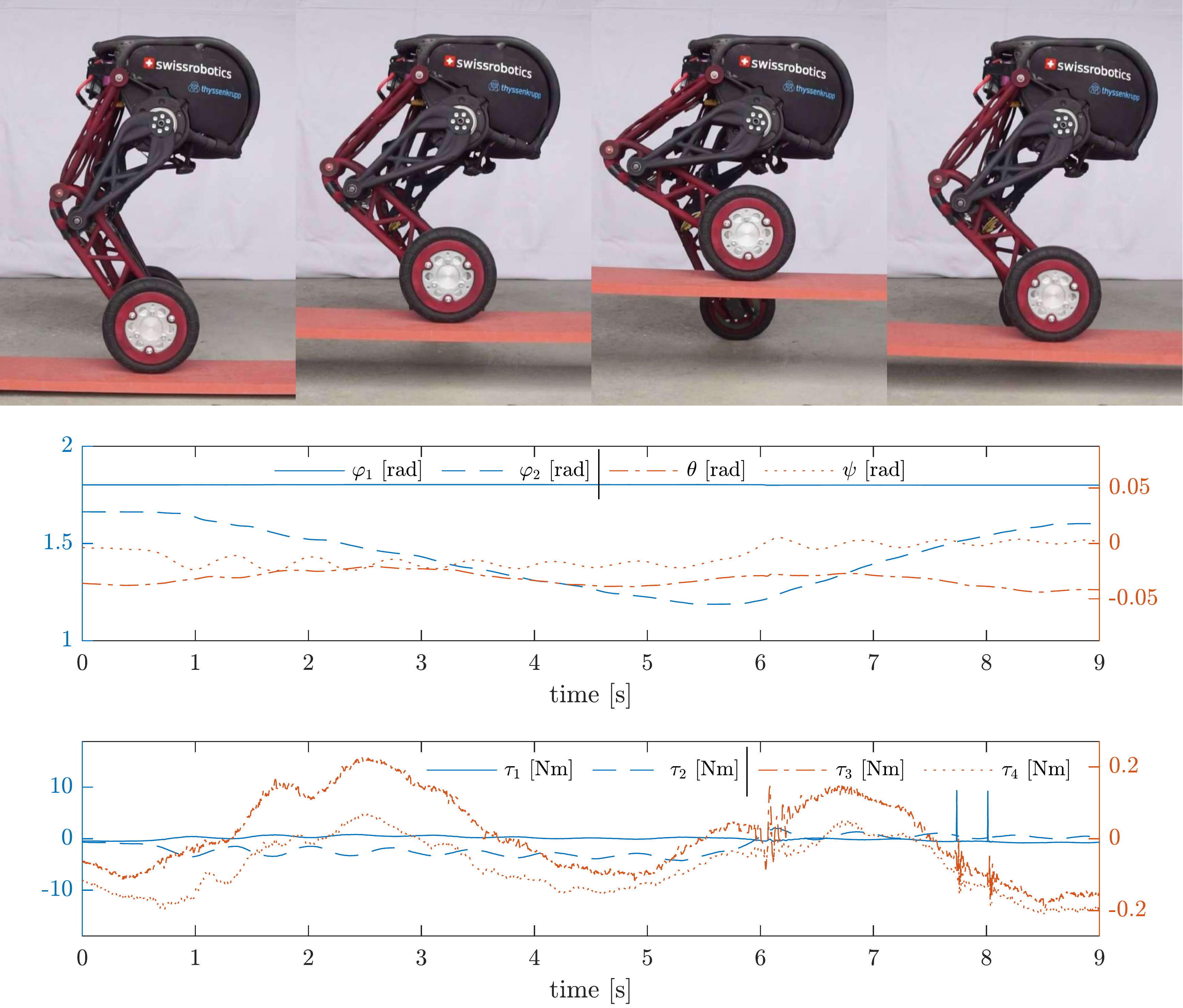}
    \vspace{-0.7cm}
	\caption{
	Adaption to varying ground heights. The robot's right wheel is placed on a plank which is \emph{manually} moved up and down, leading to unpredictable, non-periodic disturbances, as shown in the image sequence. The bottom graph shows the actuation torques computed by the \ac{WBC} and the top graph the resulting base roll angle, base pitch angle, and hip joint angles (as introduced in \figref{fig:gen_cor}). As can be seen, the robot tracks constant base height and zero roll angle by adapting the extension of the right leg accordingly.
	}
	\vspace{-0.5cm}
	\label{fig:exp_leg}
\end{figure}

The experiment shown in \figref{fig:exp_leg} demonstrates how compliance to uneven terrain arises naturally by the proposed \ac{WBC} scheme and task selection. Namely, by requiring zero roll angle, i.e. $\psi=0$, the robot remains upright by adapting the leg extensions to account for varying ground heights. This is shown while the robot is balancing and holding its position; but also while driving the exact same mechanism is active. As can be seen, the left leg extension $\varphi_1$ stays nearly constant at \SI{1.8}{rad}, while the right leg extension $\varphi_2$ closely follows the disturbance.

\subsubsection{ZMP-Regulated Curve Driving}
\label{subsubsec:curve_leaning}
\begin{figure}
	\centering
	\includegraphics[width=\columnwidth]{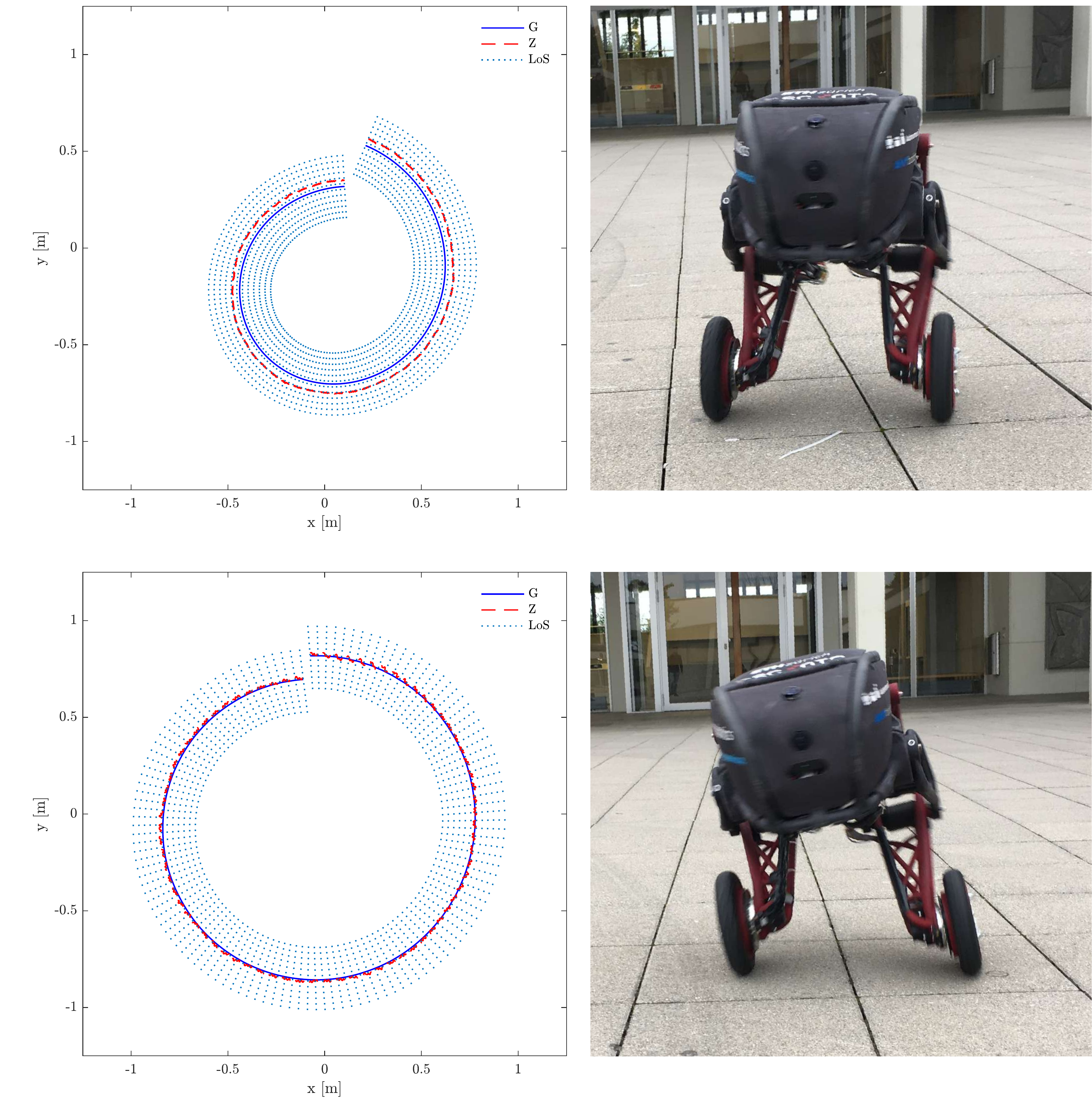}
    \vspace{-0.7cm}
	\caption{ZMP-Regulated Curve Driving.
	The graphs (with snapshots of the robot on the right) show the trajectories that result from driving a curve with constant linear (ground) and angular (yaw) velocity references, where $G$ denotes the center of the \ac{LoS} and $Z$ the \ac{ZMP}. The top graph shows the outcome from setting a constant roll angle reference of \SI{0}{\radian} and the bottom graph the one from dynamically computing the roll angle reference such as to shift the \ac{ZMP} towards $G$, which results in the robot leaning \enquote{into the curve}. 
	}
	\vspace{-0.5cm}
    \label{fig:exp_lean}
\end{figure}

To assess how computation of the roll angle reference influences curve driving, we performed two experiments. In the first one, ${}^{ref}\psi$ was set to \SI{0}{\radian} and, in the second one, it was dynamically computed to regulate the \ac{ZMP} towards the center $G$ of the \ac{LoS}, as proposed in Section~\ref{subsubsec:base_roll}. As can be seen in \figref{fig:exp_lean}, this resulted in significantly steadier curve driving when leaning. We assessed this by using the robot's wheel odometry provided by the kinematics-based state estimator.

\vspace{\sectionspace cm} \section{Outlook}
\label{sec:conclusion_and_outlook}
    In future work, we would like to extend our approach to model the dynamics of impulsive contact events, e.g. using the method outlined in~\cite{contacts}. This could for instance enable synthesis of jumping motions in a \acf{MPC} scheme leveraging the full system dynamics.
Further, we would like to improve our state estimation by incorporating measurements from additional sensors, such as cameras. Using nonlinear estimation techniques such as \ac{UKF}~\cite{wan2001unscented} or visual-inertial odometry~\cite{scaramuzza2011visual} would bring the system one step closer to autonomous deployment.
Finally, the accuracy of the dynamics model could be improved by performing dedicated system identification experiments.







\vspace{\sectionspace cm} 
\section*{ACKNOWLEDGMENT}
The authors would like to acknowledge Stefan~Kraft, Ciro~Salzmann, and Jonas~Enke for their support in real-world testing and recording of the accompanying video, as well as all supporters of the \emph{Ascento} project. Further, the first author would like to express his sincere gratitude towards Prof. Christoph Glocker for the fruitful discussions which significantly shaped the outcome of this work.


\vspace{\sectionspace cm}  
\balance
\bibliographystyle{IEEEtran}
\bibliography{IEEEabrv,submissionbibfile}

\end{document}